\documentclass{article}

\usepackage{Definitions/arxiv}  

\usepackage[utf8]{inputenc} 
\usepackage[T1]{fontenc}    

\usepackage[hyphens]{url}   
\usepackage{hyperref}       
\hypersetup{colorlinks = true,  
           	linkcolor = magenta,  
            urlcolor  = magenta,  
            citecolor = cyan,  
            anchorcolor = red}

\usepackage[sort&compress,numbers,square,comma]{natbib}  
\usepackage[colorinlistoftodos]{todonotes}

\usepackage{perpage}
\MakePerPage{footnote}  

\usepackage{doi}

\usepackage{xcolor}  

\usepackage{multirow}  
\usepackage{graphicx}  
\usepackage{tabularx}  %
\usepackage{booktabs}  

\usepackage{graphicx}  
\graphicspath{ {./images/} }

\usepackage{float}

\usepackage{caption}
\usepackage{subcaption}  
\usepackage{pgf-pie}  

\usepackage{amsfonts}       
\usepackage{amsmath}
\usepackage{amssymb}  
\usepackage{amsthm}  
\usepackage{mathtools}  
\usepackage{nicefrac}       
\usepackage{microtype}      
\usepackage{tikz}
\usetikzlibrary{matrix,calc}
\usepackage{relsize}  

\usepackage{cuted}
\usepackage{lineno}  

\usepackage{listings}  

\definecolor{tekboart_black}{RGB}{0, 0, 0}
\definecolor{tekboart_green}{RGB}{105, 177 ,38}
\definecolor{tekboart_blue}{RGB}{18, 198 ,230}
\definecolor{tekboart_orange}{RGB}{255, 128, 0}
\definecolor{tekboart_backdark}{rgb}{0.95, 0.95, 0.92}
\definecolor{tekboart_backlight}{RGB}{245, 245, 245}

\definecolor{monokai_background}{RGB}{39, 40, 34}
\definecolor{monokai_string}{RGB}{230, 219, 116}
\definecolor{monokai_comment}{RGB}{117, 113, 94}
\definecolor{monokai_text}{RGB}{248, 248, 242}
\definecolor{monokai_identifier}{RGB}{166, 226, 46}

\lstdefinestyle{izanagi}{
    numbers=left,                   		
    stepnumber=1,                   		
    numbersep=5pt,                  		
    showspaces=false,               		
    showstringspaces=false,         		
    showtabs=false,                 		
    tabsize=2,                      		
    captionpos=b,                   		
    breaklines=true,                		
    breakatwhitespace=false,         		
    keepspaces=true,
    title=\lstname,                 		
    backgroundcolor=\color{tekboart_backlight},  	
    basicstyle=\color{tekboart_black}\ttfamily\footnotesize,		
    numberstyle=\tiny\color{tekboart_black}\ttfamily,
    keywordstyle=\color{tekboart_blue}\ttfamily,	
    stringstyle=\color{tekboart_orange}\ttfamily,		
    commentstyle=\color{monokai_comment}\ttfamily,	
    emph={
        os, imghdr, pathlib,
        find, convert,
        },
    emphstyle=\color{tekboart_green}\ttfamily
}

\lstdefinestyle{monokai}{
    numbers=left,                   		
    stepnumber=1,                   		
    numbersep=5pt,                  		
    showspaces=false,               		
    showstringspaces=false,         		
    showtabs=false,                 		
    tabsize=2,                      		
    captionpos=b,                   		
    breaklines=true,                		
    breakatwhitespace=false,         		
    keepspaces=true,
    title=\lstname,                 		
    backgroundcolor=\color{monokai_background},  	
    basicstyle=\color{monokai_text}\ttfamily,		
    numberstyle=\tiny\color{monokai_black}\ttfamily,
    keywordstyle=\color{magenta}\ttfamily,	
    stringstyle=\color{monokai_string}\ttfamily,		
    commentstyle=\color{monokai_comment}\ttfamily,	
    emph={format_string, eff_ana_bf, permute, eff_ana_btr},  
    emphstyle=\color{monokai_identifier}\ttfamily
}

\title{Deep Learning for Identifying Iran's Cultural Heritage Buildings in Need of Conservation Using Image Classification and Grad-CAM\thanks{Preprint .ver 1.0: Under Submission.}}

\author{
\href{https://orcid.org/0000-0002-3320-2667}{\includegraphics[scale=0.1]{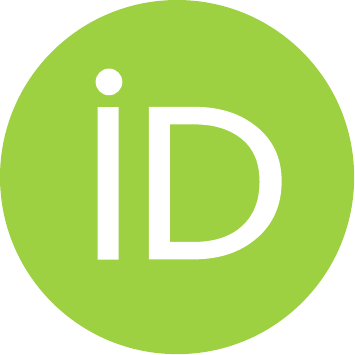}}\hspace{1mm}Mahdi~Bahrami \\
  Information Technology Department \\
  Tarbiat Modares University \\
  Tehran, Iran \\
  \texttt{mahdi\_bahrami@modares.ac.ir} \\
   \And
  Amir Albadvi \\
  Information Technology Department \\
  Tarbiat Modares University \\
  Tehran, Iran \\
  \texttt{albadvi@modares.ac.ir} \\
}

\begin{document}
\maketitle
\begin{abstract}
The cultural heritage buildings (CHB), which are part of mankind's history and identity, are in constant danger of damage or in extreme situations total destruction. That being said, it's of utmost importance to preserve them by identifying the existent, or presumptive, defects using novel methods so that renovation processes can be done in a timely manner and with higher accuracy. The main goal of this research is to use new deep learning (DL) methods in the process of preserving CHBs (situated in Iran); a goal that has been neglected especially in developing countries such as Iran, as these countries still preserve their CHBs using manual, and even archaic, methods that need direct human supervision. Having proven their effectiveness and performance when it comes to processing images, the convolutional neural networks (CNN) are a staple in computer vision (CV) literacy and this paper is not exempt. When lacking enough CHB images, training a CNN from scratch would be very difficult and prone to overfitting; that's why we opted to use a technique called transfer learning (TL) in which we used pre-trained ResNet, MobileNet, and Inception networks, for classification. Even more, the Grad-CAM was utilized to localize the defects to some extent. The final results were very favorable based on those of similar research. The final proposed model can pave the way for moving from manual to unmanned CHB conservation, hence an increase in accuracy and a decrease in human-induced errors.
\end{abstract}

\keywords{
    built cultural heritage conservation \and
    deep learning \and
    image processing \and
    convolutional neural networks (CNN) \and
    gradient weighted class activation mapping (Grad-CAM) \and
    Structural health monitoring \and
    transfer learning
}




\section{Introduction}
Two main categories of Cultural Heritage (CH) are tangible and intangible heritages, and the CHBs fall under the former category. The tangible CHs have universal values which must be physically preserved for future generations as an irreplaceable legacy \citep{Lopez2018,Vecco2010}.
CHBs are indubitably an integral part of the history and culture of human beings. Throughout the years many of these precious buildings have been in danger of damage due to several reasons, namely material deterioration, natural disasters, presence of visitors, vandalism, etc. \citep{Chen1991,Markiewicz2020,Stanco2011}.
Currently, the topic of CH has attracted increasing global attention from scientists and researchers alike, and the scope of its concept is constantly expanding. Most social scientists emphasize on its utility in supporting ethnic and national interests, while many others point to its creative and counter-hegemonic aspects \citep{Stanco2011,Brumann2015}.

\subsection{Importance}
Endowed with rich CHBs, Iran is ranked 10th in 2022, among all other countries, with 26 UNESCO world heritage sites \citep{UNESCO2022}. Although only 26 of the CHBs in Iran have been registered in UNESCO and not all of them are buildings, the number of CHBs in Iran is of the order of thousands and according to archaeological findings, Iranian architecture dates back to 6,000-8,000 B.C. \citep{Hejazi2015}. One of the reasons why Iran has been unsuccessful in registering more CHBs is the fact that most of these CHBs have not been preserved correctly, if not at all. Even some CHBs are beyond restoration.
The CHBs, which fall under the category of immovable tangible CHs, demand more sophisticated methods for conservation since we cannot move them to museums to preserve.
Lack of resources in terms of skilled practitioners, budget, and new technologies are just some of the shortcomings that introduce many problems in the conservation process. As regards the usage of state-of-the-art technologies, Iran as a developing country still uses archaic, and sometimes obsolete, manned methods to preserve these precious treasures of humanity.
From a broader perspective, many CHBs around the world suffer from such problems as well, so the use of artificial intelligence (AI) techniques such as ML and DL is not a luxury anymore but a necessity. Using ML and DL, we can move toward unmanned conservation of CHB, hence an increase in accuracy and a decrease in human-induced error.

\subsection{Research Aim}
The aim of this paper was to develop a highly generalized, yet simple, deep learning pipeline for the identification of CHBs in need of preservation, which can be used even in poor countries. We achieved this by making our model as lightweight as possible using a wealth of novel methods, as not all countries have access to expensive resources. This mindset allows for having fewer data and processing power but still reaping satisfying results (\autoref{tab:final_results}).

\subsection{Contribution}

\textbf{Unprecedented in Iran:} To the best of our knowledge, and to our surprise, not even a single scientific research had been conducted using ML or DL in the conservation of Iran's CHBs. The body of research outside Iran is not so much either. according to \citet{Fiorucci2020} the use of ML in CH literacy has been quite limited in contrast to other fields.
We believe that more research in the intersection of AI and CH can change this situation and can pave the way for the prevalence of such techniques in the process of CHB conservation around the world and accrue many benefits to CHB literacy as well.

\textbf{First-hand Complex Data: }We used first-hand data, which had been collected from different sources, as discussed in \autoref{sec:data}. Using first-hand data is important in the sense that not only our experiment would be unprecedented in Iran but globally as well; since no known CHB dataset to date \citep{Fiorucci2020} can cover the diversity of types of buildings, types of defects, and color nuances of both Persian and Islamic architecture, like ours.

\textbf{New combination of Methods: }This paper proposes an automated deep learning pipeline for identifying surface damage of CHBs. Having developing countries in mind, we used a combination of state-of-the-art methods to cater to their conservation needs with as little budget as possible. That said, the final deep learning pipeline, using a pre-trained MobileNet, can be run on low-cost devices, for instance a budget mobile phone, to make inference.

\begin{itemize}
    \item Image classification: define whether a CHB needs preservation or not.
    \item MobileNet: a very lightweight CNN architecture, but with approximately the same performance as a lot of havier CNNs (e.g., ResNet and/or Inception).
    \item Grad-CAM: to approximately localize the defects.
    \item Transfer learning: to reap great results without the need for expensive servers or manpower to take copious images.
    \item A valid data augmentation pipeline: allows the model to learn more features from the same data.
    \item Compound regularization method: a combination of four regularization methods together, namely augmentation, dropout, L2 regularization, and batch normalization.
\end{itemize}

\section{Related works}
Globally many attempts have been made to use deep learning for damage detection in CHB images.
\citet{Wang2019} used object detection (OD) with the aid of FasterR-CNN based on a ResNet101 CNN to detect damage in images of masonry buildings with bounding boxes.
In another research, \citet{Wang2020} used instance segmentation (IS), by the means of a Mask R-CNN model, for damage detection, using a masked colored layer, in glazed tiled CHBs.
An interesting work by \citet{Pathak2021} used Faster-RCNN to detect damage in CHBs, but with one major difference to other works. They used point clouds data, instead of images, as the input to their proposed model, and instead rendered point clouds as images which increased the versatility of their model, since capturing photogrammetry doesn't have the same limitations of manually taking photos.
Expectedly, damage detection using deep learning is not limited to CHB literacy; for instance, \citet{Perez2021} used OD to detect defects
on the images of modern buildings.

As highly revered as OD and IS are, they have some downsides, namely (1) a time-consuming data labeling process with bounding boxes (for OD) or color annotation (for IS); (2) the need for a huge amount of accurately labeled data; (3) detecting only pre-specified types of defects; and (4) much higher computational complexity, in comparison with image classification.
This is especially important in the case of developing countries (e.g., Iran), where budgets and resources are limited. That's why despite the prevalence of OD and IS in computer vision, many researchers opted to use the simpler image classification, where each image will be given a label as a whole, and the position of damage is not delineated.
As an example, \citet{Perez2019} used image classification and CAM layers to classify and localize defects. The downside of their work was not the use of image classification, but using cropped images, which would have been more suitable for object detection rather than image classification.

The usage of image classification and deep learning has not been just for damage detection, but aspects of CHB can benefit from them, as was the case with \citet{Llamas2017} who attempted to classify different architectural elements in historical buildings.

In terms of methodology, we followed the footsteps of \citet{Llamas2017} and \citet{Perez2019} by using image classification over OD and/or IS. Although our work is different in terms of the details of methodology and data. Unlike them, we used data augmentation and a combination of four regularization methods together, which in our case resulted in a 4-5\% improvement in metrics (\autoref{tab:final_results} and \ref{tab:compare_results}).

\textbf{Research Gap:} To the best of our knowledge, most of the works regarding deep learning and CHB use either simplistic data
or use the data belonging to a single CHB.
As a result, the final trained model lacks the generalization needed to be used for a wide range of buildings in the country of origin. We believe that the data must reflect the variety of real-world data with no editing or cropping. This way the research can come as close as possible to the practical application of using deep learning in the conservation of CHBs. Despite being known as de facto in CV, OD and/or IS need substantial computational resources to process images and detect damage, therefore making these methods infeasible for developing and/or poor countries with so many CHBs (e.g., Iran). Using more lightweight and sophisticated techniques, we can achieve reasonable results but with low-budget and simple devices (e.g., Mobile Phones).

\section{Materials and Methods}
\subsection{Data} \label{sec:data}
For this experiment, we curated a labeled dataset of approximately 10,500 CHB images. In the following, the data curation process is discussed.

\subsubsection{Data Collection}
The data were gathered from four different sources; (i) The archives of Iran's cultural heritage ministry; (ii) The author's (M.B) personal archives; (iii) images captured on site by the authors (M.B) during the research process and (iv) pictures crawled from the Internet but kept it to a minimum as their distribution differed due to heavy edits and effects. The images that didn't meet the desired quality were removed, to avoid introducing noise to our dataset.
Our collected images proved to be very challenging, in the terms of complexity, peculiarity, level of detail, and variation in size, color, characteristics, etc (\autoref{fig:data_samples}).

Regarding the population of data, as it was infeasible to have access to all the CHBs of Iran, or manually take pictures of them, we tried a random but fair approach to increase the richness of data by taking samples from a wide variety of buildings in terms of architectural style, color theme, quality, time of building, etc. In the process of collecting data different types of criteria were foremost in our minds:
\begin{itemize}
    \item \textbf{Locations}: Semnan, Hamedan, Tehran, Ghazvin, etc.
    \item \textbf{Types}: Mosques, Shrines, Churches, Palaces, etc.;
    \item \textbf{Style}: Islamic, Roman, Persian, etc.;
    \item \textbf{Types}: cracks, deterioration, mold, etc.;
    \item \textbf{Color nuances}: we have images from different times of the day and in different seasons.;
\end{itemize}

\subsubsection{Data cleaning and preprocessing} \label{sec:data_clean_preprocess}

A number of preprocessing steps were taken before creating our final datasets:
\begin{enumerate}
    \item Cleaning low-quality images, in terms of relevance, corruption, aspect ratio, grayscale, lighting condition, etc. (\autoref{fig:omitted_images}).
    \item Fixing the auto-rotation EXIF metadata.
    \item Finding a good enough resolution and resizing all images to it (i.e., 224x224).
    \item Normalizing pixel values to a range of $[-1, 1]$.
\end{enumerate}

\subsubsection{Data labeling}
Not to exacerbate the existent data imbalance, we chose binary classification over multi-class classification.
The negative class (label 0) was used for images that didn't include physical defects and the positive class (label 1) for the ones that did.

Not to become biased in the labeling phase we had three different highly qualified CHB practitioners label the images individually. This way the final label of a single image was determined by the majority vote of these three labelers.

When it comes to labeling, especially image data, we almost always have to deal with some degree of inconsistency, as different practitioners have different experiences, expertise, criteria, etc. To mitigate this effect we defined some criteria by which each labeler had a more consistent and clear guideline to label the images.
\autoref{fig:diff_size_defects} shows why it was so crucial to have some criteria that distinctly determine what should be considered a defect (e.g., in terms of length or depth).
As regards what types of physical defects were considered in the labeling process, we can enumerate the crack, mold, stain, and deterioration as the most important ones with enough samples in our dataset.

\subsubsection{Creating the datasets} \label{sec:create_dataset}
\begin{figure*}
    \centering
    \includegraphics[width=\textwidth]{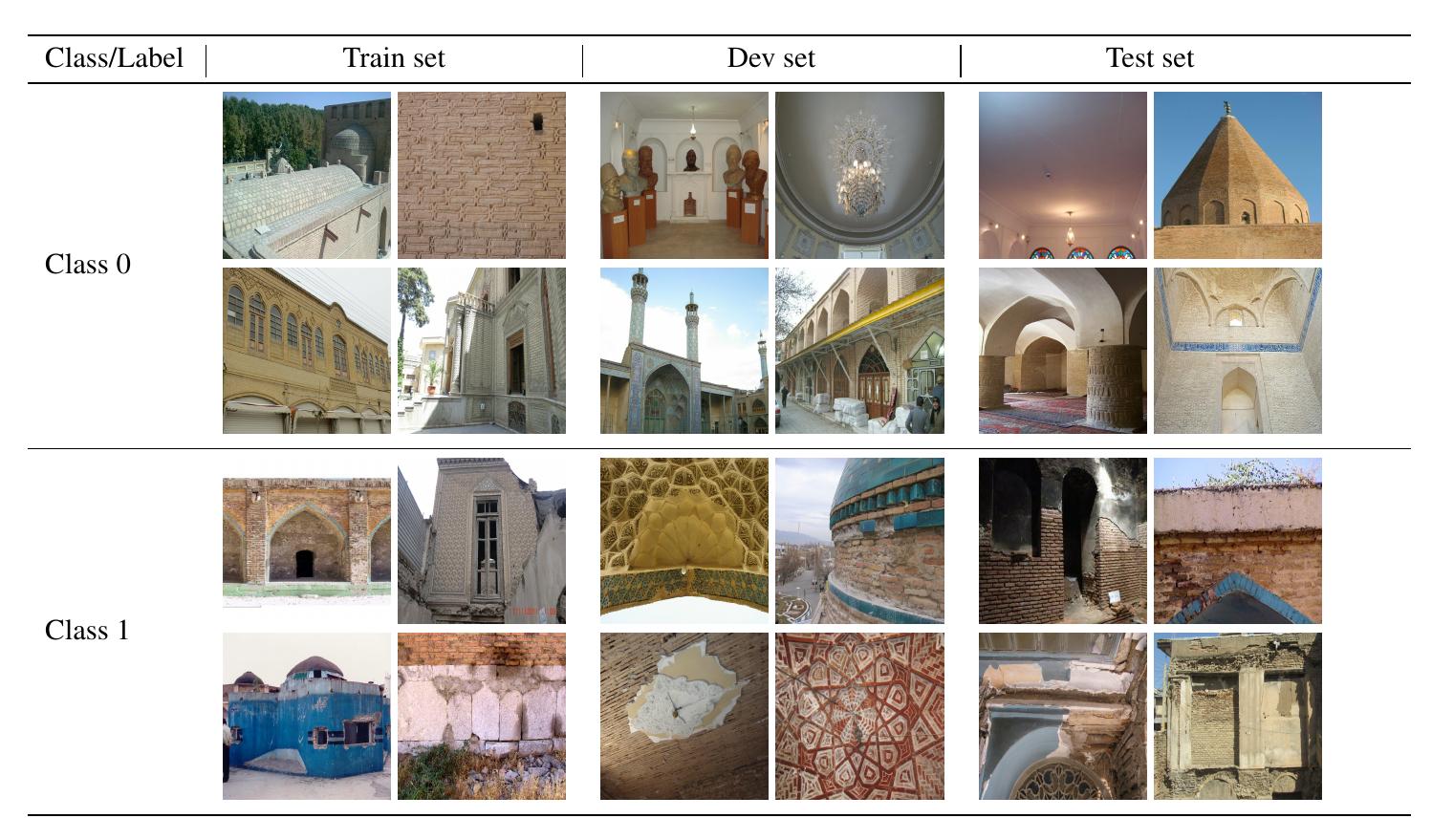}
    \caption{A few sample images which show the complexity, diversity and variation of our data.}
    \label{fig:data_samples}
\end{figure*}

After cleaning and preprocessing our data, it was divided into three mutually exclusive and jointly exhaustive sets, namely train, validation (aka dev), and test (\autoref{fig:data_samples}). To ensure a random but fair division we used stratifying shuffle that's why we have approximately the same ratio between the number images for each label (\autoref{tab:class_distri}).

\begin{table}[h]
    \centering
    \caption{The distribution of data; both aggregated and for each dataset separately.\label{tab:class_distri}}
    \resizebox{0.7\columnwidth}{!}{%
    \begin{tabular}{ccccc}
    \toprule
    \textbf{class/label} & \textbf{Total images} & \textbf{Train set}      & \textbf{Validation set} & \textbf{Test set} \\
    \midrule
    \textbf{negative/0}  & 1432         & 1018 (13.8\%)  & 207 (12.99\%)  & 207 (13.28\%) \\
    \textbf{positive/1}  & 9096         & 6358 (86.2\%)  & 1386 (87.01\%) & 1352 (86.72\%) \\
    \midrule
    \textbf{Total}       & 10528        & 7376 (70.06\%) & 1593 (15.13\%) & 1559 (14.80\%) \\
    \bottomrule
    \end{tabular}%
    }
\end{table}

As it's evident in \autoref{tab:class_distri}, the notorious yet prevalent problem of data imbalance could be identified.
A will be discussed in \autoref{sec:evaluation} we used a weighted loss function to mitigate this problem by a large margin.


\subsection{Convolutional Neural Networks (CNNs)}
Synonymous with unassailable performance when it comes to processing image data, the CNNs were a staple in the field of CV since their introduction in 1989 by LeCun et al. \citep{LeCun1989,fang2020computer}. Therefore it was somewhat indubitable that we needed to process our CHB images with this type of NNs to benefit from all the advantages that could accrue to our models by using CNNs.
Goodfellow et al. \citep{goodfellow2016deep} believe CNNs to have three main benefits: translation equivariance, sparse connections, and parameter sharing. A CNN network has less number of learnable parameters in comparison with its conventional fully connected (FC) counterpart. This reduction in the number of parameters is the product of having sparse connections and parameter sharing which enables CNNs to; (i) train faster; (ii) be less prone to overfitting and as results demand fewer train data; and (iii) be able to work with high dimensional data (e.g., images), that their FC counterparts are incapable of. The CNN does the onerous work of feature extraction automatically; the task that without CNNs used to be done by hand engineering the features \citep{gu2018recent}.

In this experiment, we used three of the most prestigious CNN architectures which have shown compelling results and interesting loss convergence, namely ResNet \citep{He2015}, Inception \citep{Szegedy2014}, and MobileNet \citep{Howard2017}.

\subsection{Transfer Learning} \label{sec:transfer_learning}
Dealing with several restraints such as lack of enough data and powerful computers, a methodology called transfer learning was employed to drastically mitigate these impediments. TL tries to transfer the knowledge, a pre-trained model has already learned from a large amount of data, to another model \citep{Zhuang2019}.
Generally, TL consists of two main parts. The first part is responsible for customizing the output layer to our problem. The second part fine-tunes the pre-trained model to adapt more to our specific data.

\subsection{Class Activation Mapping (CAM)}
In spite of the merits of image classification, there is a notorious drawback that lies within, and that is the black-box nature of artificial neural networks (NN). That being said, we don't know whether the model considers pertinent features in an image to decide its class or not. That's why researchers came up with a solution named class activation mapping (CAM) \citep{Zhou2015}.

In this experiment we used gradient-weighted class activation maps (Grad-CAM) \citep{Selvaraju2016} which is a CAM method that merges the gradients (aka derivatives) of the final classification, that is the output layer deciding the label of the image, and the output of the final Conv layer of the model to generate a heatmap. The heatmap then is applied to the original image to localize the places that were taken into account when deciding its class/label.

\subsection{Regularization} \label{sec:regularization}
\begin{figure}[h]
    \centering
    \includegraphics[width=0.7\columnwidth]{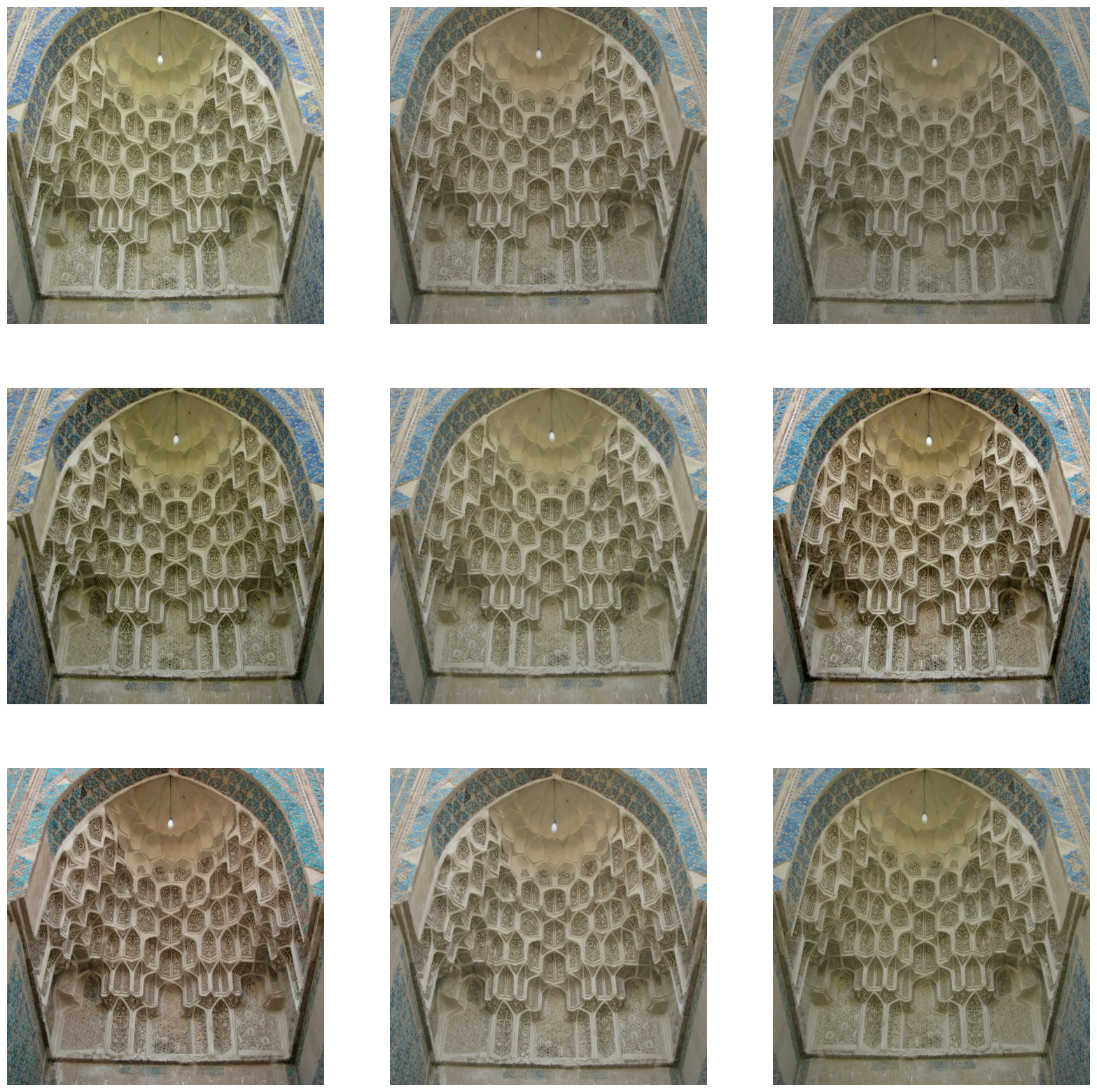}
    \caption{An example of applying the proposed data augmentation methods on a train image (i.e., nine times). Notice how random, realistic, and valid the augmented versions are.} \label{fig:data_aug}
\end{figure}

As one of the salient reasons for the occurrence of overfitting is the lack of enough data, which is ubiquitous in CV, we are always in need of more data. Unfortunately getting more brand-new data is not always possible. A workaround is to use the data we already have to increase the number of valid labeled train data, hence a decrease in overfitting as the model is now less capable of naively memorizing the train set \citep{Maharana2022}.
As data augmentation is a staple in CV \citep{Maharana2022}, we almost always opt for using it and this paper is not exempt. Finally, in \autoref{fig:data_aug} the result of our proposed data augmentation pipeline after nine runs on the same image can be seen. The data augmentation methods used in this paper can be found in \autoref{tab:data_aug}.

\begin{table}[h]
    \centering
    \caption{The data augmentation methods used in this paper and their corresponding values.\label{tab:data_aug}}
    \resizebox{0.5\columnwidth}{!}{%
    \begin{tabular}{cc|cc}
    \toprule
    \textbf{method}   & \textbf{value} & \textbf{method}     & \textbf{value} \\
    \midrule
    random flip     & Horizontal     & random brightness & 0.05           \\
    random rotation & 0.005          & random saturation & 0.6 - 1.2      \\
    random crop     & 5\%            & random contrast   & 0.75 - 1.1     \\
    random quality  & 80 - 100       & random hue        & 0.03           \\
    \bottomrule
    \end{tabular}%
    }
\end{table}

Briefly, to decrease overfitting, which is commonplace in DL models, due to their high capacity in terms of the number of parameters, a combination of four famous methods were used, namely L2 regularization \citep{Cortes2012}, dropout \citep{Hinton2012}, batch normalization layer \citep{Ioffe2015}, and data augmentation \citep{Maharana2022}. The results of this combining approach, as discussed in \autoref{sec:results}, were quite satisfiable in terms of overfitting and resulted in a very small amount of overfitting (i.e., $< 1\%$) for all of our models.

\section{Implementation}
\subsection{Network Architecture}
\begin{figure*}[h]
\centering
\includegraphics[width=\textwidth]{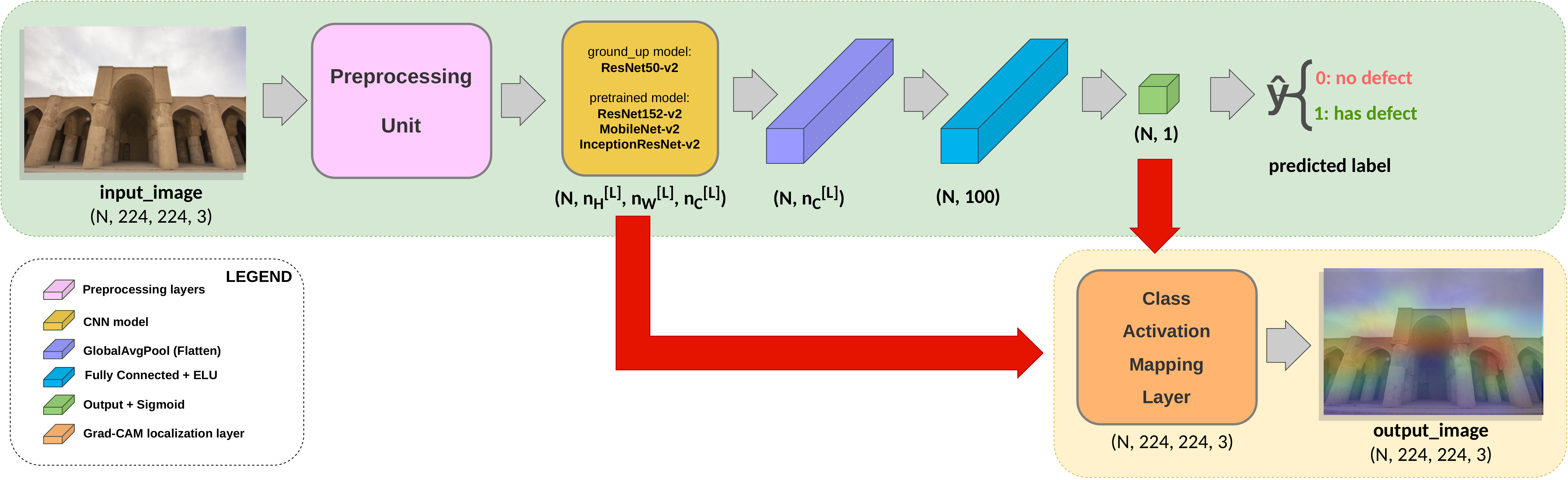}
\caption{The overall architecture of our proposed model/network. Where the values shown in parenthesis below each layer represent the layer's output shape. The $N$, $n_H^{[L]}$, $n_W^{[L]}$, and $n_C^{[L]}$ refer to the batch size, height, width, and channels of the last layer ($L$) of the embedded CNN model respectively. \label{fig:arch}}
\end{figure*}

In the \autoref{fig:arch} the holistic architecture of our proposed method is represented.
Not to process new input images through a data preprocessing pipeline every time, we embedded both the resizing and the normalization preprocessing functions into our network (i.e., pink box). This way, there would be no need to process the unknown images before executing the prediction on them, after the model had been trained.

It was alluded to before that in this experiment we made use of several pre-eminent CNN architectures to tackle the problem at hand and not to be biased toward a certain architecture. As a result, four different networks were implemented, namely ResNet50-v2, ResNet152-v2, InceptionResNet-v2, and MobileNet-v2. One main difference between the ResNet50-v2 and other models is that we trained the ResNet50-v2 from scratch and with randomly initialized weights; while the other three were pre-trained models which were accompanied by TL.

The responsibility of the Global Average Pooling layer (i.e., purple box) was to flatten the output of the last Conv layer into a matrix, which is the desired shape of the input of a fully connected (FC) layer.
Before replacing the output of the pre-trained model with a layer of our own, an FC layer (i.e., light blue box) was added to decrease underfitting; the bigger our network becomes the less underfitting we experience, but it also increasing overfitting, that's why a single FC layer proved to provide a desired trade-off, and thus reduced underfitting by a large margin without increasing overfitting too much.

As shown in \autoref{fig:arch}, our model has two outputs. The first (i.e., green box) is responsible for the task of classification, by which each image will be given a label (i.e., negative or positive). The second output on the other hand does the task of localizing the parts by which the model has decided on a certain label for a specific image; this task is done by the Grad-CAM method (i.e., orange box).

\subsection{Evaluation} \label{sec:evaluation}
To evaluate the implemented networks several metrics have been used in an endeavor to meticulously monitor the behavior of the networks at different stages of training. All these metrics will be scrutinized in the following subsections.

\subsubsection{Cost function}
As mentioned in \autoref{sec:create_dataset} our two classes were imbalanced and since it would nudge our model to be biased toward the class with more examples (i.e., the positive class), we had to tackle this problem somehow. Having decided in favor of using the class weight method due to its numerous merits the \autoref{eq:calc_class_weights} was used to calculate the weight of each class, but it's worth noting that there is a myriad of ways to calculate the weights but as we would fine-tune the calculated weights later on in hyperparameter tuning phase we chose the most widely used:

\begin{equation} \label{eq:calc_class_weights}
    \mathlarger{\displaystyle w_c = \frac{n_t}{n_l * n_c}}
\end{equation}

Where $w_c$, $n_t$, $n_l$, and $n_c$ indicate the calculated weight of class $c$, the total number of images in the dataset, the number of classes, and the number of images in class $c$ respectively. These weights then will be used in the cost function of our networks so that the importance of images belonging to the inferior class outweighs that of the superior class, in a way that network will be rewarded or penalized more when it comes to the images of the class with fewer examples in it. The binary cross-entropy cost function was used, and the way it calculates cost before and after applying class weights can be seen in \autoref{eq:cost_without_class_weight} and \ref{eq:cost_with_class_weight} respectively. To make it more concrete the first one is used in validation, test, and prediction while the latter is employed in training time; that is we only care about data imbalance during training which is common sense as the network only updates its internal parameters (e.g., weights) in training time and backpropagation.

\begin{equation} \label{eq:cost_without_class_weight}
L(\hat{y}, y) = -\bigg(ylog(\hat{y}) + (1 - y)log(1 - \hat{y})\bigg)
\end{equation}

\begin{equation} \label{eq:cost_with_class_weight}
L(\hat{y}, y) = -\bigg((w_1)(y)log(\hat{y}) + (w_0)(1 - y)log(1 - \hat{y})\bigg)
\end{equation}

Where $y$ refers to the true label and the $\hat{y}$ to the predicted label of the given record. Note that as we did binary classification and sigmoid activation function for the output layer then $\hat{y}$ is actually the probability ([0, 1]) of the record belonging to the positive class.

\subsubsection{Performance measures and metrics} \label{sec:metrics}
When it comes to the evaluation of our model, several metrics were incorporated to ensure the rigor of our results. As we suffer from imbalanced data the Accuracy can be quite misleading if the model gets biased toward the superior class, so to address this issue four more performance measures were used, namely Precision, Recall, F-Score, and AUC. If anything, the F-Score is the harmonic mean of the Precision and Recall, thus it takes into account both of them to give us a balanced score of the two. Mathematically, Accuracy, Precision and Recall, and F-Score are defined as:

\begin{equation} \label{eq:accuracy}
Accuracy = {\frac{TP + TN}{TP + FP + TN + FN}}
\end{equation}

\begin{equation} \label{eq:precision}
    Precision = {\frac{TP}{TP + FP}}
\end{equation}

\begin{equation} \label{eq:recall}
    Recall = {\frac{TP}{TP + FN}}
\end{equation}

\begin{equation} \label{eq:f1}
    F\!\!-\!\!Score = {\frac{2 * Precision * Recall}{Precision + Recall}}
\end{equation}

Where TP, TN, FP, and FN are True Positive, True Negative, False Positive, and False Negative respectively. In this paper the FN takes precedence over FP, thus the Recall is more important than precision as the FN is in the denominator of the Recall's \autoref{eq:recall}, however, we tried to balance them as much as possible. The reason is that if an image is falsely labeled as positive then in the worst-case scenario we lose time, but in the case of an image being falsely labeled as negative, then a building in dire need of conservation can be overlooked which might lead to irredeemable destruction. The area under the ROC curve, abbreviated as AUC, was employed in an endeavor to refrain from creating a model biased toward a certain class. AUC demonstrates the power of the model in distinguishing different classes.

\section{Results} \label{sec:results}
After slogging through the onerous task of training and fine-tuning the hyperparameters several times, we achieved highly satisfactory results (\autoref{tab:final_results}). Note that the training process of the ResNet50-v2 doesn't have the fine-tuning step as we trained it from the ground up and with random initial weights. Considering the lack of enough data and computational power, which were alluded to before, it was of no surprise that the networks trained with TL fared the best.

Among the networks that used TL, there is no definite winner, but the MobileNet-v2 had the best performance considering both the performance measures and the computational complexity for both the training and making an inference.
That said, MobileNet's lightweight architecture is conducive to training and predicting faster which is especially important for devices with low computational power such as mobile phones, edge devices, etc. which are considered de facto pieces of equipment to monitor CHBs \citep{Maksimovic2019}.

\begin{table*}[h]
\centering
\caption{Final results, after hyperparameter tuning.\label{tab:final_results}}
\resizebox{.85\textwidth}{!}{%
\begin{tabular}{cccc|ccc|ccc|ccc}
\toprule
\multirow{2}{*}{\textbf{Measure}}            & \multicolumn{3}{c}{\textbf{ResNet50V2 \textsuperscript{1}}}      & \multicolumn{3}{c}{\textbf{ResNet152V2 \textsuperscript{2}}}     & \multicolumn{3}{c}{\textbf{MobileNetV2 \textsuperscript{2, 3}}}     & \multicolumn{3}{c}{\textbf{InceptionResNetV2 \textsuperscript{2}}} \\
\cmidrule{2-13}
                                             & \textbf{train} & \textbf{val} & \textbf{test} & \textbf{train} & \textbf{val} & \textbf{test} & \textbf{train} & \textbf{val} & \textbf{test} & \textbf{train}  & \textbf{val}  & \textbf{test} \\
\midrule
\textbf{Loss}                                & 0.48           & 0.47         & 0.48          & 0.38           & 0.38         & 0.38          & 0.31           & 0.32         & 0.33          & 0.36            & 0.36          & 0.37          \\
\textbf{Accuracy}                            & 0.83           & 0.84         & 0.83          & 0.88           & 0.89         & 0.89          & 0.90           & 0.90         & 0.90          & 0.88            & 0.88          & 0.88          \\
\textbf{Precision}                           & 0.87           & 0.87         & 0.87          & 0.92           & 0.92         & 0.92          & 0.95           & 0.94         & 0.94          & 0.91            & 0.91          & 0.91          \\
\textbf{Recall}                              & 0.95           & 0.95         & 0.95          & 0.95           & 0.95         & 0.96          & 0.94           & 0.94         & 0.94          & 0.96            & 0.95          & 0.96          \\
\textbf{F-Score}                           & 0.91           & 0.91         & 0.91          & 0.93           & 0.94         & 0.94          & 0.94           & 0.94         & 0.94          & 0.93            & 0.93          & 0.93          \\
\textbf{AUC}                                 & 0.54           & 0.54         & 0.54          & 0.89           & 0.88         & 0.88          & 0.93           & 0.92         & 0.90          & 0.87            & 0.86          & 0.85          \\
\textbf{TP}                                  & 6040           & 1310         & 1287          & 6056           & 1319         & 1296          & 5961           & 1311         & 1274          & 6082            & 1319          & 1295          \\
\textbf{FP}                                  & 923            & 189          & 192           & 551            & 107          & 114           & 328            & 78           & 76            & 623             & 123           & 135           \\
\textbf{TN}                                  & 95             & 21           & 22            & 467            & 103          & 100           & 690            & 127          & 139           & 395             & 87            & 79            \\
\textbf{FN}                                  & 318            & 73           & 67            & 302            & 64           & 302           & 397            & 77           & 79            & 276             & 64            & 59            \\
\bottomrule
\end{tabular}%
}\\
\begin{flushleft}
    \noindent{\footnotesize{\textsuperscript{1} This NN was trained from scratch and with initial random weights.}}\\
    \noindent{\footnotesize{\textsuperscript{2} These NNs were trained and fine-tuned using TL.}}\\
    \noindent{\footnotesize{\textsuperscript{3} This NN has the best performance among all.}}
\end{flushleft}
\end{table*}

\subsection{Evaluation of MobileNet-v2's Performance}
As mentioned before and according to \autoref{tab:final_results} the fine-tuned model made with pre-trained MobileNet-v2 was the winner among the other three networks. and its lightweight architecture which is conducive to training and predicting faster is especially important for devices with low computational power such as mobile phones, edge devices, etc. That being said, as the winner among all four network architectures let's scrutinize MobileNet-v2's performance even more. The results of other networks in detail can be found in \autoref{fig:appendix:resnet50v2_PM}-\ref{fig:appendix:inception_PM}. The \autoref{tab:hyperparameters} displays the most important hyperparameters used during the training and fine-tuning of our multiple networks.

\begin{figure*}[h]
    \centering
    \includegraphics[width=0.9\textwidth, trim={0 0 0 6.5cm}, clip]{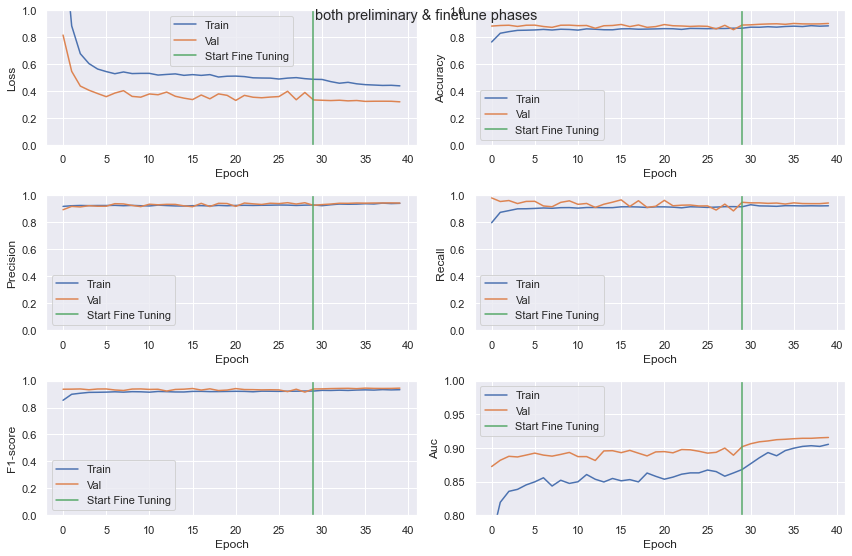}
    \caption{The changes in performance measures reported after each epoch for both the train and validation sets during the training and fine-tuning phase; belonging to the MobileNet-v2 network. the green line indicates the point, in terms of epoch number, where we started to fine-tune some late layers in the pre-trained model.}
    \label{fig:MobieNetV2_metrics}
\end{figure*}

\begin{figure*}[h]
    \centering
    \includegraphics[width=0.8\textwidth]{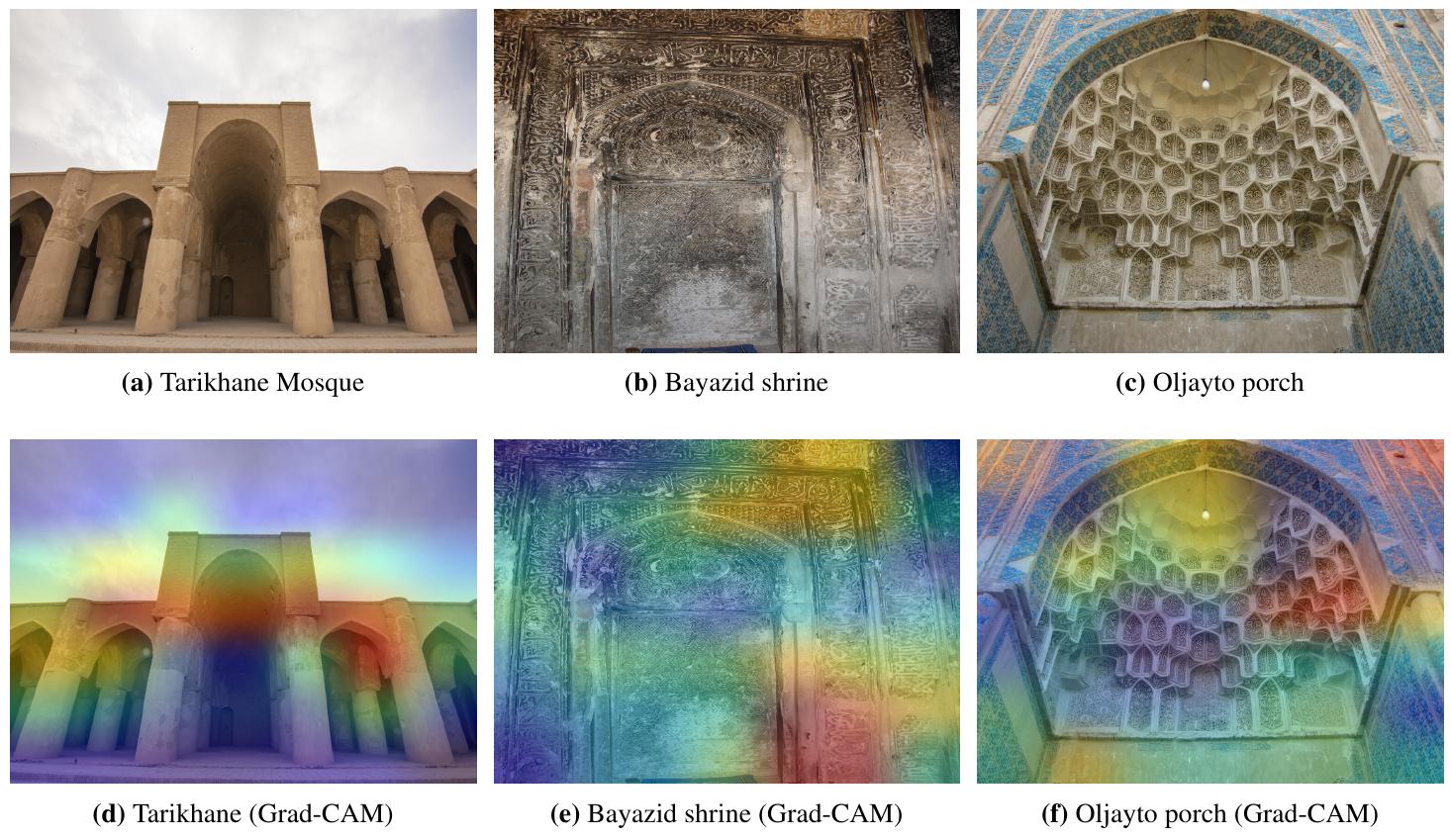}
    \caption{Some samples of the output of Grad-CAM layer of fine-tuned MobileNet-v2 network. The localized defects are shown by a heatmap (from Blue to Red). \label{fig:cam_layer}}
\end{figure*}

The fine-tuned MobileNet-v2 doesn't suffer from underfitting nor overfitting (\autoref{fig:MobieNetV2_metrics}).
As regards the second output of the fine-tuned MobileNet-v2, the localizations seemed completely relevant and attest to the fact that the model had learned the correct features in the train data (\autoref{fig:cam_layer}).

\begin{figure*}[h]
    \centering
    \includegraphics[width=0.9\textwidth]{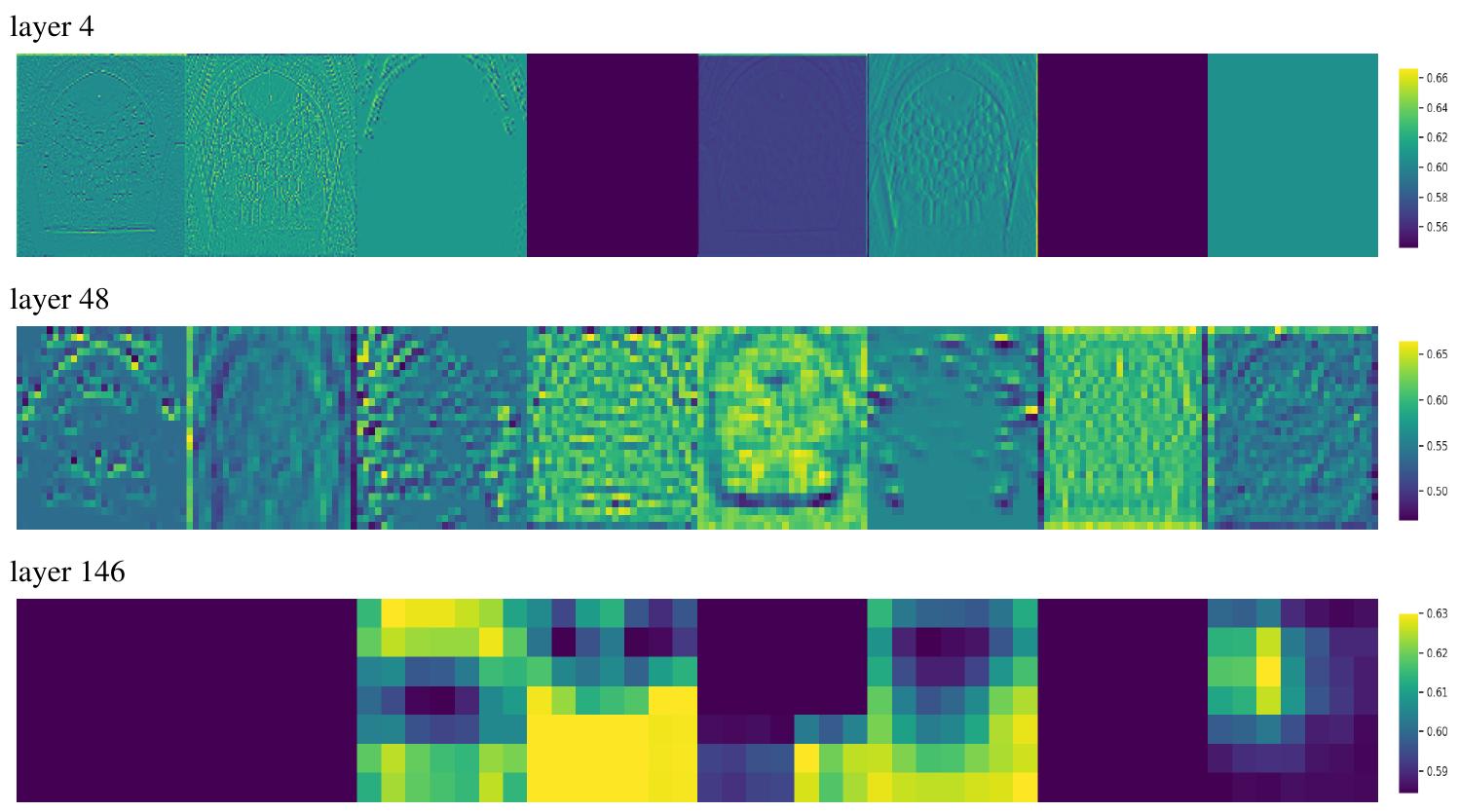}
    \caption{A few samples (i.e., 8) of feature maps from the beginning (top), mid-section (middle), and end (bottom) of our fine-tuned MobileNet-v2 network. The input image was the same as that of the Of subfigure~c~in~\autoref{fig:cam_layer}. \label{fig:feature_maps}}
\end{figure*}

The output of several Conv layers, aka feature maps, from our fine-tuned MobileNet-v2 network, are visualized in \autoref{fig:feature_maps}; we purposefully chose one layer from the beginning, one from the middle, and another from the end of the network to demonstrate that the more we go deep into the network the more holistic and abstract the detected features will be and vice versa.

\section{Discussion}
This work demonstrates the facilities of DL in the conservation of CHB by the means of damage detection. As we have collected a diverse set of intricate CHB images, the trained model is very robust and achieved a minimum of 90\% for all the metrics we used on the test set. More than our diverse data, using TL, data augmentation, and three different regularization methods in combination, was conducive to reducing overfitting and increasing the generalization power of our model.
The salient reasons that attest to why our results are considered to be good enough are (i) Bayes error rate and (ii) the value of performance measures. Although measuring Bayers error rate is a hard and time-consuming task, which was not in the scope of this experiment, we can argue that its value is high, as for instance even a highly skilled CHB practitioner from the south of Iran, would have had a hard time detecting the defects in CHBs from north of the country, considering the peculiarity and idiosyncrasies of each building in our dataset.

According to \citet{Mandrekar2010}, in the field of CV, values larger than 90\% are considered excellent, so it's safe to assume that the MobileNet-v2 had excellent performance, recording values above 90\% for all of our metrics. Other than reaching the best performance among other models, the MobileNet-v2 is particularly interesting as it is a faster NN which is particularly important in doing real-time damage detection in devices with low computational resources, such as mobile phones or edge devices. Using our proposed model based on MobileNet-v2 can pave the way for the wide usage of such models in CH sites in Iran and/or around the world with the fewest possible resources.

To compare our results with those of similar researchers, the papers of \citet{Llamas2017} and \citet{Perez2019} were used, as these were the ones that used image classification, CNN, and TL, just like this experiment. As both of these papers used multiclass classification whereas we used binary classification, we took the average of each metric (e.g., Recall) for all classes, Llamas et al. had ten/10 classes and Perez et al. had four/4 classes; this way we could make their results comparable to those of ours. The comparison of the results on the test set is shown in \autoref{tab:compare_results}.

\begin{table}[h]
\centering
\caption{A comparison between the results of similar studies. the reported values are for test set.\label{tab:compare_results}}
\resizebox{0.6\columnwidth}{!}{%
    \begin{tabular}{lccc}
    \toprule
    {}    & Precision & Recall & F1 Score \\
    \midrule
    Llamas et al. (ResNet) \protect\citep{Llamas2017} & 0.90              & 0.90             & 0.90          \\
    Perez et al. (VGG-16) \protect\citep{Perez2019}   & 0.90              & 0.89             & 0.89          \\
    Our fine-tuned model (MobileNet-v2)               & \textbf{0.94}     & \textbf{0.94}    & \textbf{0.94} \\
    \bottomrule
    \end{tabular}%
}
\end{table}

The most important challenges and limitations that we faced during this experiment were: (i) needing more data, which is a perennial problem in CV; (ii) lack of suitable computational power; and (iii) inconsistency in labeling due to personal preference and difference in the level of labelers' expertise.

\section{Conclusion}
This experiment is concerned with applying novel yet matured methods such as DL and CNNs to make the process of conservation of CHBs less prone to errors and more efficient than doing it manually by direct human supervision. By getting Iran's CHB practitioners, the main beneficiaries of this experiment, to use our proposed models besides their old methods, a higher rate of success in detecting physical defects of such buildings can be achieved. We irrevocably believe that CHB practitioners using DL models, such as our proposed one, can identify physical defects more often than either does alone and hopefully as a result, a lower prospect of CHBs deteriorating in structural health.

In an endeavor to practically demonstrate the utilities of DL in CH literature, We developed a fully fledged DL model that classifies the images in need of conservation and even more approximately localizes the defects to help the CH practitioners identify defects in a timely manner, and as a result speed of the process of CHB conservation as well as increasing its accuracy.
In spite of all the limitations, we achieved very good results with a score of at least 94\% for Precision, Recall, and F1-Score, which were about 4-5\% more than similar works (\autoref{tab:compare_results}).

As regards future works, addressing the limitations we faced can open up a plethora of opportunities in terms of methods and outputs. for instance, if had access to a large amount of labeled data and powerful servers, physical or in the cloud, then object detection or instance segmentation would be more useful and could elicit more accurate and user-friendly results from our data. Having gotten traction in the past few years, the generative adversarial networks (GANs) can be utilized in our network architecture to propose restoration based on the label and localizations our proposed model offers.

\nolinenumbers  
\bibliography{Definitions/refs}  

%
%
%
%

\newpage
\appendix  
\numberwithin{equation}{section}
\numberwithin{figure}{section}
\numberwithin{table}{section}
\numberwithin{lstlisting}{section}

\section{Appendix: Supplementary Materials} \label{appendix:full_results}

\begin{figure*}[h]
    \centering
    \includegraphics[width=0.79\textwidth]{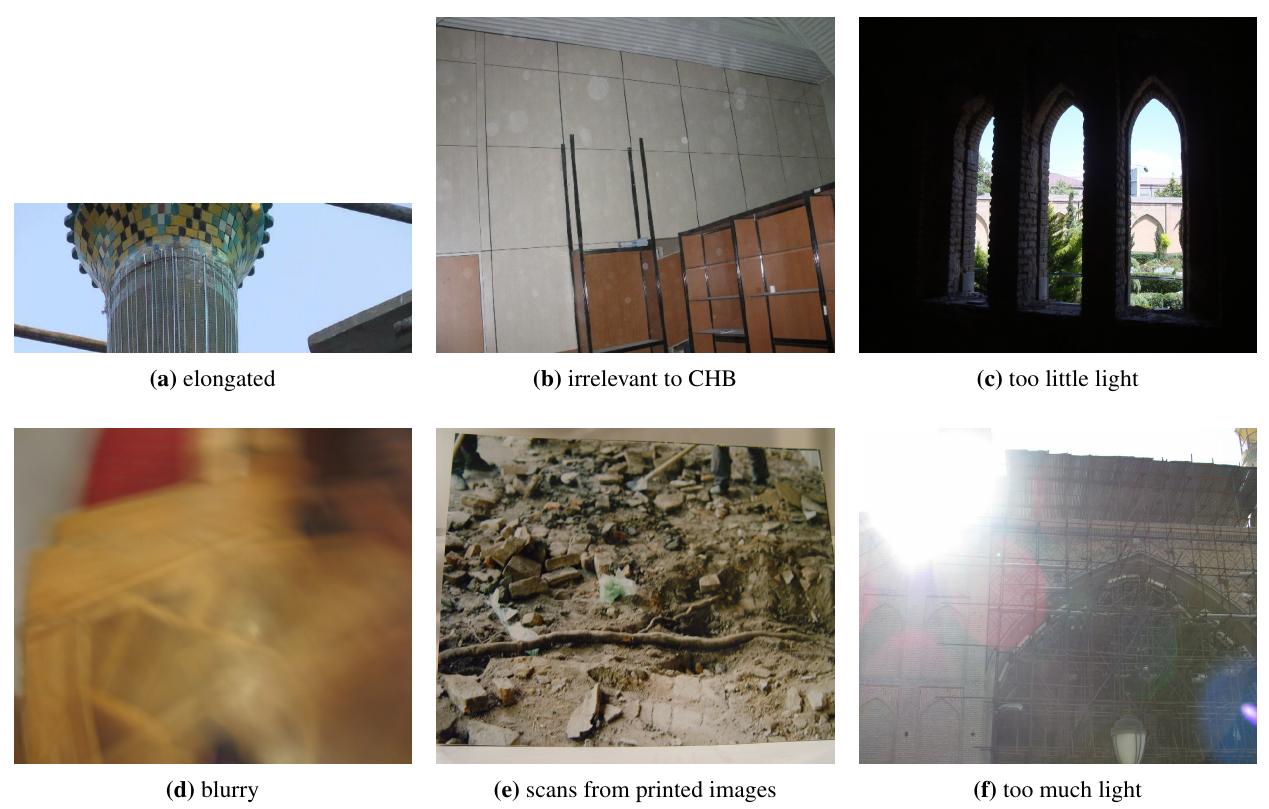}
    \caption{Some examples of unsuitable images, which were omitted in the data preprocessing phase. \label{fig:omitted_images}}
\end{figure*}

\begin{figure}[h]
    \centering
    \includegraphics[width=0.6\textwidth]{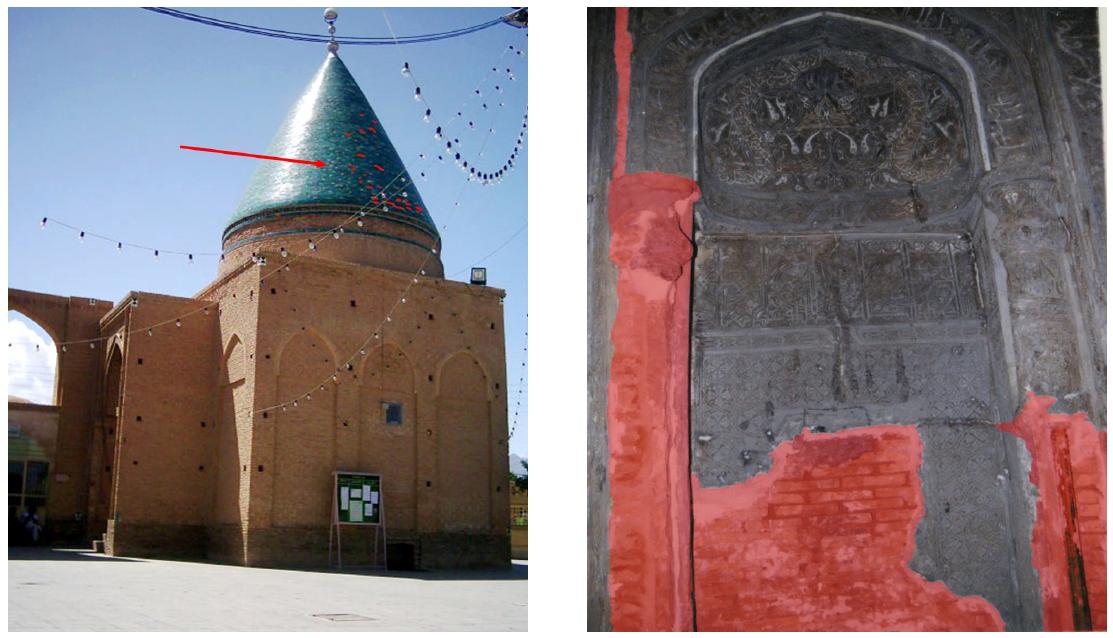}
    \caption{Comparing a picture with small-sized defects (left) with a picture with large-sized defects (right). The defects are delineated in red, for more clarity.\label{fig:diff_size_defects}}
\end{figure}

\begin{figure*}[h]
    \centering
    \includegraphics[width=0.68\textwidth]{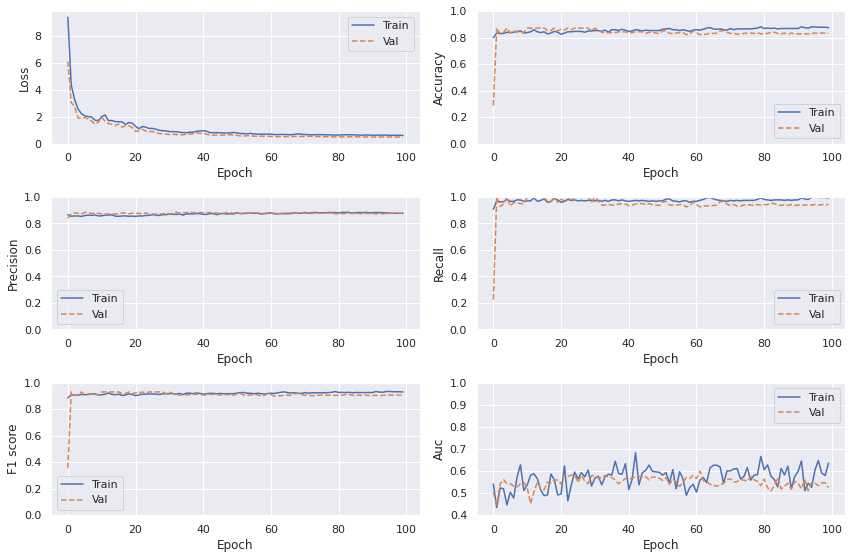}
    \caption{The changes in performance measures reported after each epoch for both the train and validation error for the ResNet50-v2 network.\label{fig:appendix:resnet50v2_PM}}
\end{figure*}

\begin{figure*}[h]
    \centering
    \includegraphics[width=0.68\textwidth, trim={0 0 0 1cm}, clip]{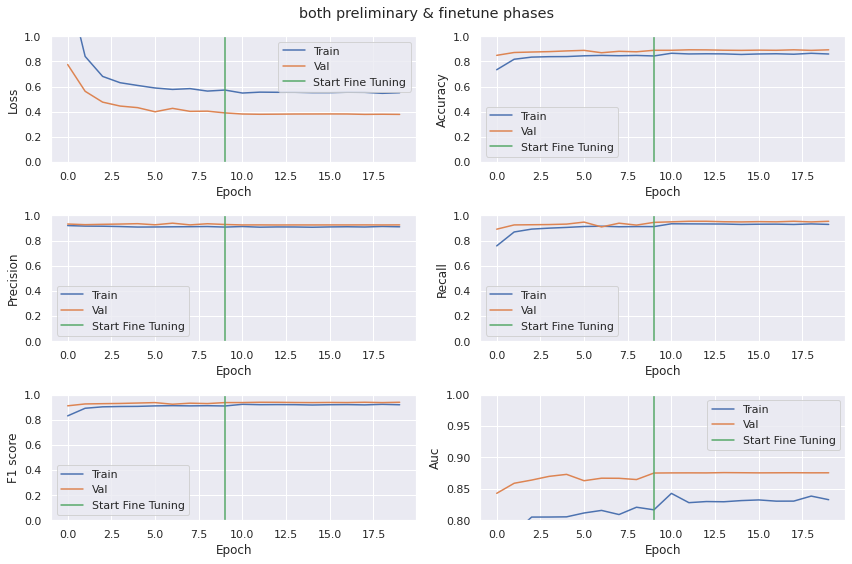}
    \caption{The changes in performance measures reported after each epoch for both the train and validation error belonging to the ResNet152-v2 network.\label{fig:appendix:resnet152v2_PM}}
\end{figure*}

\begin{figure*}[h]
    \centering
    \includegraphics[width=0.68\textwidth, trim={0 0 0 1cm}, clip]{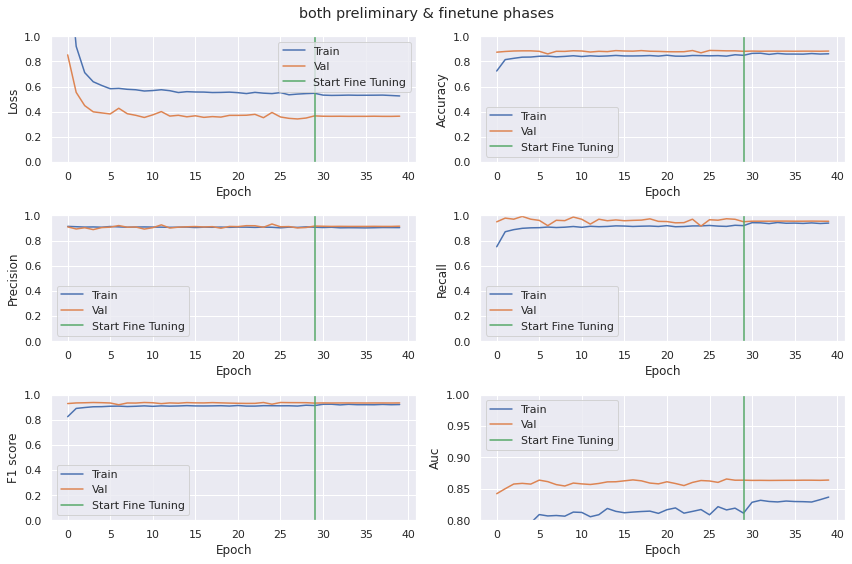}
    \caption{The changes in performance measures reported after each epoch for both the train and validation error belonging to the InceptionResNet-v2 network.\label{fig:appendix:inception_PM}}
\end{figure*}

\begin{table*}[h]
\caption{The salient hyperparameters used to train our networks.\label{tab:hyperparameters}}
\begin{tabularx}{\textwidth}{lccccc}
\toprule
\textbf{hyperparameter}      & \textbf{ResNet50V2} & \textbf{ResNet152V2} & \textbf{MobileNetV2} & \textbf{InceptionResNetV2} & \textbf{default} \\
\midrule
\textbf{batch\_size}         & 32                   & 32                    & 32                    & 32                          & -                      \\
\textbf{feature range}       & {[}-1, 1{]}          & {[}-1, 1{]}           & {[}-1, 1{]}           & {[}-1, 1{]}                 & {[}0, 255{]}           \\
\textbf{dropout}             & 0.8                  & 0.7                   & 0.5                   & 0.5                         & 0.0                    \\
\textbf{L2 lambda}           & 0.1                  & 0.01                  & 0.01                  & 0.01                        & 0.01                   \\
\textbf{class weight}        & \{'0': 3.5, '1': 1\} & \{'0': 2, '1': 1\}    & \{'0': 2, '1': 1\}    & \{'0': 3, '1': 1\}          & \{'0': 1, '1': 1\}     \\
\textbf{optimizer}           & Adam                 & Adam                  & Adam                  & Adam                        & RMSprop                \\
\midrule
first step \textsuperscript{1}\\
\textbf{learning rate}       & 0.01                 & 0.005                 & 0.001                 & 0.001                       & 0.001                  \\
\textbf{decay steps}         & 1,000                & 1,000                 & 1,000                 & 1,000                       & 100,000                \\
\textbf{decay rate}          & 0.96                 & 0.96                  & 0.96                  & 0.96                        & 0.96                   \\
\textbf{\#epochs}            & 100                  & 10                    & 30                    & 30                          & -                      \\
\midrule
second step \textsuperscript{2}\\
\textbf{learning rate}       & -                    & 1e-8                  & 1e-6                  & 1e-8                        & 0.001                  \\
\textbf{decay steps}         & -                    & 300                   & 300                   & 300                         & 100,000                \\
\textbf{decay rate}          & -                    & 0.96                  & 0.96                  & 0.96                        & 0.96                   \\
\textbf{\#epochs}            & -                    & 10                    & 10                    & 10                          & -                      \\
\textbf{\#unlocked layers \textsuperscript{3}} & -  & 64 out of 564         & 54 out of 154         & 80 out of 780               & -                      \\
\bottomrule
\end{tabularx}%
\vspace{0.1cm}
\noindent{\footnotesize{\textsuperscript{1} Refers to the first part of TL where we train the weights of the added FC layers at the end of pre-trained models.}}\\
\noindent{\footnotesize{\textsuperscript{2} Refers to the second part of TL where we fine-tune the weights of several filters in the pre-trained model.}}\\
\noindent{\footnotesize{\textsuperscript{3} This value shows how many of the later layers in the pre-trained models were unlocked to be fine-tuned in second step of TL.}}\\
\end{table*}

\end{document}